\documentclass[letterpaper]{article}
\usepackage{conf}  
\usepackage{times}  
\usepackage{helvet}  
\usepackage{courier}  
\usepackage{url}  
\usepackage{graphicx}  
\usepackage{subcaption}
\usepackage{amsmath,amssymb}
\usepackage{algorithm}
\usepackage{algpseudocode}
\usepackage{multirow}
\usepackage{booktabs}
\newtheorem{definition}{Definition}
\newcommand{\schemename}{TREP}

\begin{document}
\title{
	Representation Learning of Pedestrian Trajectories Using\\
    Actor-Critic Sequence-to-Sequence Autoencoder
}
\author{
	Ka-Ho Chow, Anish Hiranandani, Yifeng Zhang, S.-H. Gary Chan\\
	Department of Computer Science and Engineering\\
	Hong Kong University of Science and Technology\\
	khchowad@cse.ust.hk, ahiranandani@connect.ust.hk, yzhangch@cse.ust.hk, gchan@cse.ust.hk
}
\maketitle
\begin{abstract}
Representation learning of pedestrian trajectories transforms variable-length timestamp-coordinate tuples of a trajectory into a fixed-length vector representation that summarizes spatiotemporal characteristics. It is a crucial technique to connect feature-based data mining with trajectory data. Trajectory representation is a challenging problem, because both environmental constraints (e.g., wall partitions) and temporal user dynamics should be meticulously considered and accounted for. Furthermore, traditional sequence-to-sequence autoencoders using maximum log-likelihood often require dataset covering all the possible spatiotemporal characteristics to perform well. This is infeasible or impractical in reality. We propose \schemename{}, a practical pedestrian \underline{\textbf{T}}rajectory \underline{\textbf{REP}}resentation learning algorithm which captures the environmental constraints and the pedestrian dynamics without the need of any training dataset. By formulating a sequence-to-sequence autoencoder with a spatial-aware objective function under the paradigm of actor-critic reinforcement learning, \schemename{} intelligently encodes spatiotemporal characteristics of trajectories with the capability of handling diverse trajectory patterns. Extensive experiments on both synthetic and real datasets validate the high fidelity of \schemename{} to represent trajectories.
\end{abstract}

\section{Introduction}
With  advances in both indoor and outdoor positioning technologies and  penetration of  mobile devices, there has been a proliferation of location-aware applications for navigation, gaming, recommendations, social meetups, etc. These applications collect large volume of ordered sequences of user locations~\cite{vatsavai2012spatiotemporal}. These data, termed \emph{pedestrian trajectories}, contain rich information describing user behaviors.  Trajectory mining is to extract these behaviors and understand user profiles, so that novel or more personalized services can be offered.

\begin{figure}
	\centering
	\includegraphics[width=\columnwidth]{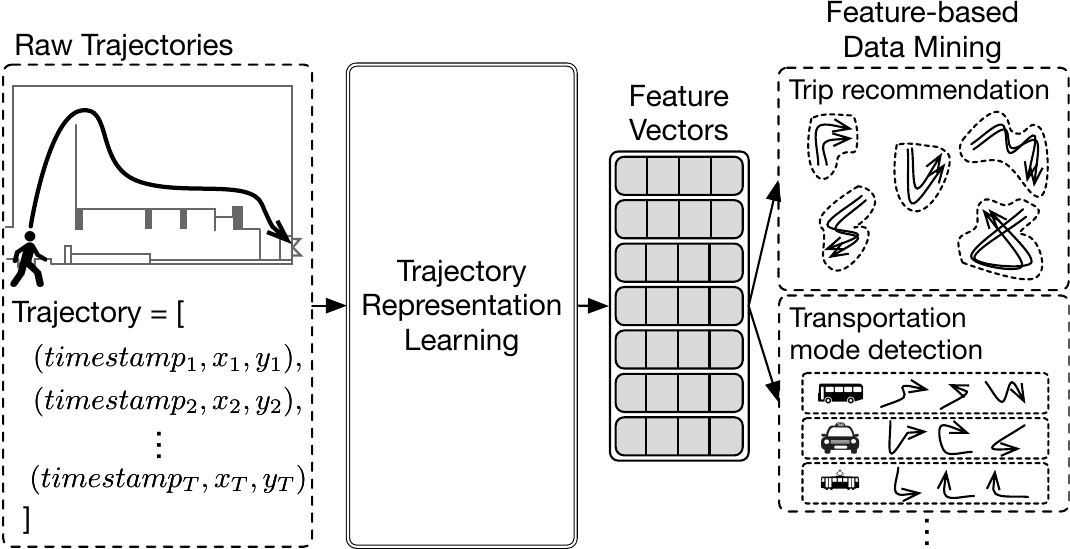}
	\caption{Trajectory representation learning serves as an adapter to transform timestamp-coordinate tuples of a user trajectory into a feature vector for feature-based data mining.}
	\label{fig:intro}
\end{figure}

Trajectory mining, despite its opportunities, is a challenging problem. This is because a trajectory is often represented as an ordered sequence of timestamp-coordinate tuples, but many algorithms widely used for data analytics (such as support vector machines) require input data to be in vector space. To bridge this gap, we need representation learning, a machine learning approach to represent a trajectory as a feature vector that captures  the spatial characteristics of the locations and the dynamic pathways of the pedestrians.

Representation learning has long been an important field in machine learning. It transforms different kinds of structured (e.g., a graph) or unstructured (e.g., a corpus) data into a vector representation that can still preserve the unique features in their original form~\cite{bengio2013representation}. In this paper, we study representation learning for trajectory data.

Figure~\ref{fig:intro} illustrates the role of trajectory representation learning in the process  of knowledge discovery. It takes trajectories, in the form of timestamp-coordinate tuples, as input and generates feature vectors that compactly represent them.  The feature vectors are then used as input to drive data mining applications. For example, the preferences of a mall visitor can be inferred by analyzing the types and the duration of the shops he/she visits. Through user profiling and clustering~\cite{ying2011semantic}, trip recommendations or personalized advertisements can be made to increase the revenue of the mall and improve the customer experience. In addition, transportation modes can be classified from the representation of GPS trajectories~\cite{stenneth2011transportation}. They contain invaluable information for urban planners when designing an intelligent transportation system that resolves traffic congestion and improves the quality of life.

Unlike vehicular trajectories extensively investigated before~\cite{zheng2015trajectory}, representation learning for {\em pedestrian} trajectories is distinct and far more challenging. This is because vehicle dynamics is rather homogeneous due to the hard constraints imposed by the road conditions (e.g., traveling direction and speed). These conditions often become the principles to handcraft the representations~\cite{jiang2017trajectorynet}. In contrast, pedestrian movements are semi-constrained as they may roam freely along different pathways. The dynamics is hence much more heterogeneous and random, and little motion characteristics can be exploited to derive representations in a similar manner.  In other words, the previous approaches developed for vehicular trajectory representation cannot be extended to our context of pedestrian trajectory.

Besides exploiting motion characteristics as in vehicular trajectories, another typical approach is to use an autoencoder. Such approach uses artificial neural networks to compress the data (e.g., an image) into a representation which  can reconstruct the original input through decompression. By extending it to a sequence-to-sequence architecture~\cite{srivastava2015unsupervised}, one can easily handle the temporal information hidden in the sequence and convert it into a vector representation. Nevertheless, such approach has not considered the spatial characteristics such as the constraints introduced by walls or forbidden/infeasible regions. Furthermore, training a sequence-to-sequence autoencoder using the traditional maximum log-likelihood often suffers from overfitting, because it memorizes the training samples without taking into account of the environmental constraints and  pedestrian dynamics. Moreover,  the model cannot handle trajectories with rather diverse patterns,  because this requires a full dataset covering all the possible spatiotemporal characteristics.

To overcome the above challenges, we propose \schemename{}, a novel pedestrian {\textbf{T}}rajectory {\textbf{REP}}resentation learning algorithm based on a sequence-to-sequence autoencoder with a spatial-aware objective function. It models trajectories by learning the action sequences  (e.g., moving forward, turning right, etc.) taken by the pedestrians. Due to the non-differentiable objective function and the absence or incompleteness of training data, we propose to optimize \schemename{} in an actor-critic reinforcement learning fashion. We reformulate the spatial-aware objective function as a reward function in reinforcement learning. By interacting with the environment through explorations and exploitations, \schemename{} learns to generate trajectory representations that contain sufficient information to produce policies maximizing the reward. Based on our extensive experiments on both synthetic and real datasets, \schemename{} exhibits  superior understanding of pedestrian dynamics in a semi-constrained walkway and the capability of handling trajectories with diverse patterns.

The remainder of this paper is organized as follows. We first discuss the preliminaries and the problem statement. We then present the proposed \schemename{} framework including its neural network design and learning paradigm. We finally present illustrative experimental results before concluding the paper.

\section{Preliminaries and Problem Statement}
To simplify computations, we employ a grid abstraction on the map by partitioning it into a grid of contiguous square cells of size $\alpha$. The set of cells constitutes a new coordinate system at an abstract level as any coordinates in the original system can be mapped to a cell enclosing it.
\begin{definition}
	(actions) The set of actions $\mathcal{A}=(a_{N},a_{NE},a_{E},a_{SE},a_{S},a_{SW},a_{W},a_{NW},a_{*})$ represents the movements in eight cardinal and inter-cardinal directions, and an option $a_{*}$ to remain stationary.
\end{definition}
\begin{definition}
	(movement operator) We define $\odot$ as the movement operator. The expression $v\odot a$ returns the cell reached by taking an action $a$ at cell $v$.
\end{definition}
\begin{definition}
	(reachable cells) The set of reachable cells from a cell $v$, denoted by $RCells(v)$, is defined as the set of cells a pedestrian can reach physically by taking an action at $v$ under environmental constraints.
\end{definition}
\begin{figure}
	\centering
	\includegraphics[width=0.62\columnwidth]{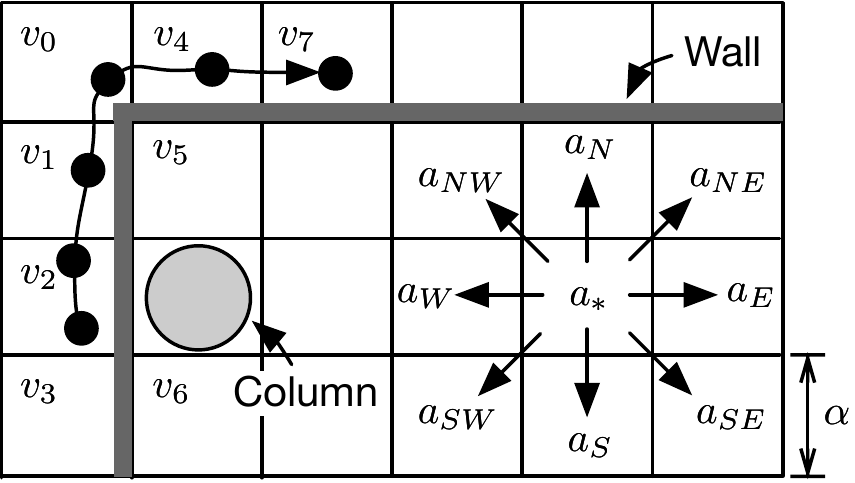}
	\caption{An indoor space with environmental constraints.}
	\label{fig:preliminary}
\end{figure}
Based on the above, the pedestrian road network is modeled by a directed graph $G=(\mathcal{V}, \mathcal{E})$ where $\mathcal{V}=\{v_0, v_1, \dots, v_N\}$ is a set of $N$ cells satisfying environmental constraints and $\mathcal{E}=\{(v_i, v_j)\vert v_j\in RCells(v_i), v_i\in\mathcal{V}\}$ is a set of directed edges indicating the legal transitions between cells. Figure~\ref{fig:preliminary} depicts a pedestrian road network in an indoor space (e.g., a shopping mall) with a wall partition and a column. Due to the obstacle (i.e., the column), the cell with a gray circle is not included in $\mathcal{V}$. In this example, $v_1\odot a_N=v_0$ and $RCell(v_1)=\{v_0, v_1, v_2, v_4\}$. Note that $v_5\notin RCell(v_1)$ as the wall partition makes the movement from $v_1$ to $v_5$ by taking the action $a_E$ illegal.

\begin{figure*}[ht!]
	\centering
	\includegraphics[width=0.7\linewidth]{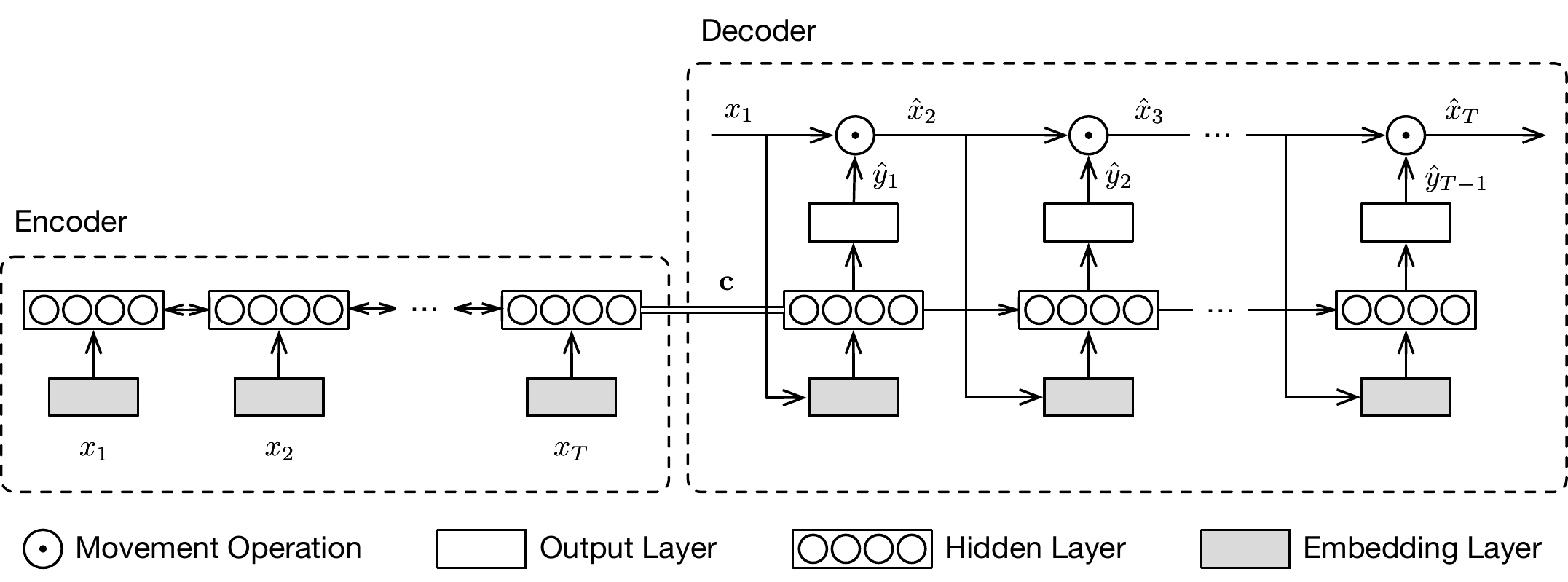}
	\caption{Illustration of the trajectory autoencoder (the actor network).}
	\label{fig:actor-network}
\end{figure*}

We consider the case where the localization system (e.g., vision-based tracking) generates high-quality trajectories. Otherwise, map matching and interpolation techniques~\cite{xiao2014lightweight} are applied, such that the sampling intervals are short and regular. Then, the trajectories can be transformed into paths on the pedestrian road network. 
\begin{definition}
(trajectory) A trajectory $X$ on a pedestrian road network $G$ is an ordered sequence of cells and is defined as: $X=(x_1, x_2, \dots, x_T)$ where $x_t\in\mathcal{V}$ and $(x_t, x_{t+1})\in\mathcal{E}$.
\end{definition}
For instance, the trajectory shown in Figure~\ref{fig:preliminary} can be represented by the path $(v_2, v_2, v_1, v_0, v_4, v_7)$.

To simplify the formulas, we ignore the fact that trajectories may be of different lengths and always use $T$ to denote the length of a trajectory. The notation $X_{s\dots e}$ denotes sub-trajectories of the form $(x_s, \dots, x_e)$. Instead of using one-hot vectors to represent cells, we conduct graph representation learning~\cite{grover2016node2vec} on the pedestrian road network to obtain their embeddings. As a cell is equivalent to a location point at an abstract level, the terms ``cell" and ``location point" are used interchangeably.

We consider the problem of learning to represent a variable-length trajectory $X$ as a fixed-length vector representation $\mathbf{c}\in\mathbb{R}^D$. Given a set of trajectories, their representations can be used as input to feature-based data mining.

\section{\schemename{} Framework}
In this section, we first discuss the architecture of \schemename{}. Then, the learning paradigm based on actor-critic reinforcement learning is presented.

\subsection{Trajectory Autoencoder}
A recurrent neural network (RNN) is a neural network which can process a sequence $(x_1, \dots, x_T)$ with an arbitrary length $T$. It reads tokens one by one and produces a sequence of hidden state vectors $(h_1,\dots, h_T)$, where $h_t$ at time $t$ is updated by a transition function $f: h_t=f(h_{t-1}, x_t)$. The hidden state vector $h_t$ is a summary of the tokens seen by the RNN as, by recursively unfolding $h_t=f(f(f(\dots), x_{t-1}), x_t)$, it is a function of the sub-sequence $X_{1\dots t}$. The transition function $f$ is usually equipped with a gating mechanism to alleviate the vanishing and exploding gradient problems. A popular choice is the Long Short-Term Memory (LSTM)~\cite{hochreiter1997long}.

To predict a class label $y_t$ at time $t$ based on the sub-sequence $X_{1\dots t}$, a stochastic output layer is applied on the hidden state vector. A probability distribution $p(\cdot\vert X_{1\dots t})$ conditioned on the input tokens will be generated, and the label with the maximum probability will be selected. In this paper, we leverage RNNs with LSTM to model trajectories by referring the above terms ``sequence" and ``token" to ``trajectory" and ``location point" respectively.

\schemename{} is a trajectory autoencoder with an encoder-decoder architecture~\cite{sutskever2014sequence}, which is depicted in Figure~\ref{fig:actor-network}. The encoder is a bidirectional RNN~\cite{schuster1997bidirectional} that reads location points $(x_1,\dots,x_T)$ in both forward and backward directions, producing two sequences of hidden state vectors $(h^F_1,\dots,h^F_T)$ and $(h^B_1,\dots,h^B_T)$. The last hidden state vectors (i.e., $h^F_T$ and $h^B_T$) are concatenated to form the fixed-length vector representation $\mathbf{c}$ that summarizes the entire trajectory.

The decoder is a unidirectional RNN taking a summary $\mathbf{c}$ from the encoder and a starting location point $x_1$ as inputs. Unlike the RNN in the encoder, the hidden state vector $h_t$ in the decoder is also conditioned on the summary $\mathbf{c}$:
\begin{equation}
h_t=f(h_{t-1}, \hat{x}_{t}, \mathbf{c}).
\end{equation}
At each time step $t$, the stochastic output layer computes the conditional distribution of the action $\hat{y}_t$ taken by the pedestrian located at $\hat{x}_t$:
\begin{equation}
p(\cdot\vert\hat{X}_{1\dots t}, \mathbf{c})=\frac{\exp(Wh_t+b)\otimes \mathcal{M}_{\hat{x}_t}}{\vert\vert\exp(Wh_t+b)\otimes \mathcal{M}_{\hat{x}_t}\vert\vert_1},
\end{equation}
where $W$ and $b$ are the parameters of the output layer, $\otimes$ is the element-wise multiplication and $\mathcal{M}_{v_i}\in\mathbb{R}^{\vert\mathcal{A}\vert}$ is a mask vector with
$$\mathcal{M}_{v_i,a}=
\begin{cases}
1 & \text{if } v_i\odot a\in RCells(v_i), \\
0 & \text{otherwise.}
\end{cases}
$$
The above state-constrained softmax~\cite{wu2017modeling} can effectively guarantee that the action taken is restricted by the environmental constraints. The pedestrian will take action $\hat{y}_t$ and move from $\hat{x}_{t}$ to $\hat{x}_{t+1}$, which becomes the input at time $t+1$.

The encoder and decoder are jointly trained. The encoder extracts spatiotemporal features from the variable-length trajectory and encodes them into a fixed-length vector representation $\mathbf{c}$, which provides sufficient information for the decoder to generate a sequence of actions $\hat{Y}=(\hat{y}_1,\dots,\hat{y}_{T-1})$, such that the reconstructed trajectory $\hat{X}=(\hat{x}_1,\dots,\hat{x}_T)$, where $\hat{x}_{t+1}=\hat{x}_t\odot\hat{y}_t$, resembles the ground-truth $X$. The fixed-length vector representation $\mathbf{c}$ is the output of \schemename{} and can be used in data mining.

\begin{figure*}
	\centering
	\includegraphics[width=0.7\linewidth]{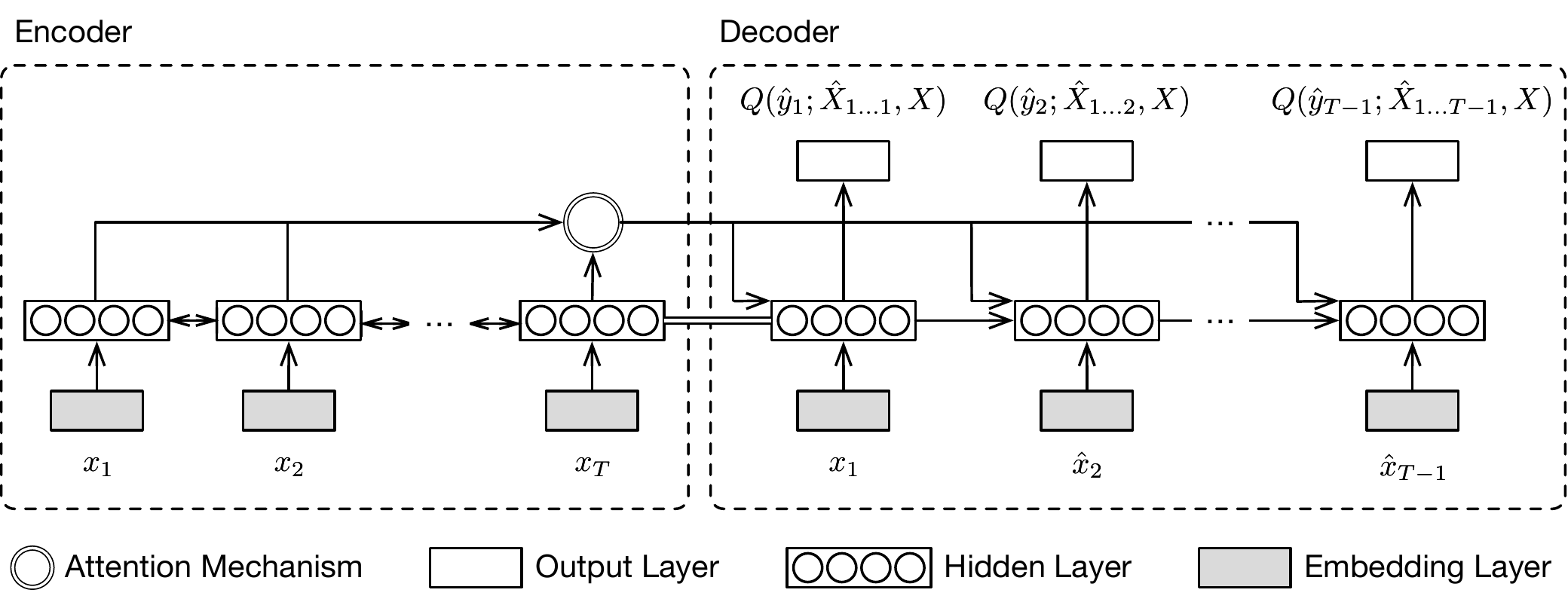}
	\caption{Illustration of the $Q$-value estimator (the critic network).}
	\label{fig:critic-network}
\end{figure*}

\subsection{Actor-Critic Reinforcement Learning Paradigm}
One way to train the trajectory autoencoder is to maximize the log-likelihood. Given the ground-truth sequence of actions $Y$, the log-likelihood $\mathcal{L}_{ll}$ can be maximized by updating the trajectory autoencoder, parameterized by $\phi$, with the following gradient:
\begin{equation}\label{eq:logll}
\frac{d\mathcal{L}_{ll}}{d\phi}=\frac{d}{d\phi}\bigg(\sum_{t=1}^{T-1}\log p(y_t\vert\hat{X}_{1\dots t}, \mathbf{c})\bigg).
\end{equation}
However, there are at least two drawbacks in using this learning objective:
\begin{itemize}
\item
{\em Ignoring the spatial dimension}:
The neural network memorizes the training sample pairs $(X, Y)$ without exploiting the spatial dimension, even if the RNN can take care of the temporal one. We argue that to better model pedestrian trajectories, it should be learned from criticizing the reconstructed trajectory $\hat{X}$ in terms of the error compared with the ground-truth $X$ on the pedestrian road network $G$. Concretely, the trajectory autoencoder should be learned to minimize the spatial-aware objective function:
\begin{equation}\label{eq:hl-obj}
\mathcal{D}(X, \hat{X})=\sum_{t=2}^{T}d(x_t, \hat{x}_t),
\end{equation}
where $d(\cdot,\cdot)$ is a shortest path distance measure on $G$.
\item
{\em Insufficient training data and weak learning signal:}
Collecting a full training dataset that contains all the possible spatiotemporal characteristics is infeasible. Even if a massive dataset can be obtained, the ground-truth action $y_t$ is valid only if the preceding predictions are correct, i.e., $(y_1,\dots,y_{t-1})=(\hat{y}_1,\dots,\hat{y}_{t-1})$, making the learning signal weak especially when the trajectories are lengthy.
\end{itemize}
In light of the above, we propose to optimize \schemename{} using the spatial-aware objective function (i.e., Equation~\ref{eq:hl-obj}). Due to the non-differentiable nature of the shortest path distance function and the absence of training data, we reformulate the problem in a reinforcement learning setting. The experiments validate that minimizing Equation~\ref{eq:hl-obj} is the key to leading \schemename{} to a superior performance in handling trajectories with diverse patterns.

To formulate a reinforcement learning problem, we define the total reward $\mathcal{R}(X,\hat{X})$ of a reconstructed trajectory $\hat{X}$ given the ground-truth $X$ as
\begin{equation}\label{eq:reward}
\mathcal{R}(X,\hat{X})=\sum_{t=1}^{T-1}r(\hat{y}_t;\hat{X}_{1\dots t},X).
\end{equation}
The immediate reward received when the pedestrian moves to $\hat{x}_{t+1}$ by taking the action $\hat{y}_t$ at $\hat{x}_t$ is given by
\begin{equation}
r(\hat{y}_t;\hat{X}_{1\dots t},X)=
\begin{cases}
0 & \text{if } d(x_{t+1}, \hat{x}_{t+1})\leq\delta, \\
-1 & \text{otherwise,}
\end{cases}
\end{equation}
where $\delta$ is the tolerable step-wise reconstruction error. For example, $\delta=0$ means the reconstructed trajectory must be exactly the same as the ground-truth to get a non-negative reward.

Recall in reinforcement learning that the value function represents the expected future reward in a particular state of the system. In our case, the state of the system is the location of the pedestrian reached by taking a sequence of predicted actions starting at $x_1$. The value of the pedestrian at $\hat{x}_t$ is defined as
\begin{equation}\label{eq:v-func}
\begin{aligned}
V(\hat{x}_t; X)=&V(\hat{Y}_{1\dots t-1}; X)\\
=&\displaystyle\mathop{\mathbb{E}}_{\hat{Y}_{t\dots T-1}\sim p(\cdot\vert\hat{X}_{1\dots t}, \mathbf{c})}\sum_{\tau=t}^{T-1}r(\hat{y}_\tau;\hat{X}_{1\dots \tau}, X).
\end{aligned}
\end{equation}
Intuitively, the value is the expected number of time steps that the pedestrian can move along the ground-truth within an error bound $\delta$ in the future.

We aim at training the trajectory autoencoder in such a way that the sequence of predicted actions $(\hat{y}_1,\dots,\hat{y}_{T-1})$ maximizes the total reward (Equation~\ref{eq:reward}) or, equivalently, minimizes the spatial-aware objective function (Equation~\ref{eq:hl-obj}). It is also the same as maximizing the value (Equation~\ref{eq:v-func}) in a stochastic setting. The $Q$-value of a candidate action $a$ is the expected future reward after performing the action $a$ at the current location and can be computed by
\begin{equation}
\begin{aligned}
&Q(a;\hat{X}_{1\dots t},X)\\
=&\displaystyle \mathop{\mathbb{E}}_{\hat{Y}_{t+1,\dots,T-1}\sim p(\cdot\vert\hat{X}_{1\dots t+1},\mathbf{c})}\bigg(r(a;\hat{X}_{1\dots t},X)+\\
&\sum_{\tau=t+1}^{T-1}r(\hat{y}_\tau;\hat{X}_{1\dots \tau}, X)\bigg).
\end{aligned}
\end{equation}
Then, the trajectory autoencoder can be trained using the gradient of the expected value~\cite{bahdanau2016actor}:
\begin{equation}\label{eq:actor-grad}
\frac{dV}{d\phi}=\displaystyle \mathop{\mathbb{E}}_{\hat{Y}\sim p(\cdot\vert x_1,\mathbf{c})}\frac{d}{d\phi}\bigg(\sum_{t=1}^{T-1}\sum_{a\in\mathcal{A}}p(a\vert\hat{X}_{1\dots t},\mathbf{c})Q(a;\hat{X}_{1\dots t},X)\bigg),
\end{equation}
which is also known as the actor-critic. An actor (i.e., the trajectory autoencoder) predicts an action $\hat{y}_t$ by generating a policy $p(\cdot\vert\hat{X}_{1\dots t},\mathbf{c})$. A critic criticizes it through the expected future reward $Q(\cdot;\hat{X}_{1\dots t},X)$ and improves the subsequent predictions by increasing the probability of the actions that give high $Q$-values, and decreasing the probability of the actions that give low  $Q$-values.

We estimate the $Q$-values in Equation~\ref{eq:actor-grad} parametrically by using a critic network (Figure~\ref{fig:critic-network}), another neural network with an encoder-decoder architecture. The encoder takes the ground-truth trajectory as input and produces a sequence of hidden state vectors, which are fed to a soft attention mechanism~\cite{bahdanau2014neural}. The output layer of the decoder is designed with a dueling network architecture~\cite{wang2015dueling} to estimate the $Q$-values.

To train the critic network, the Bellman equation is utilized to compute the target $Q$-value $q_t$ of the action $\hat{y}_t$:
\begin{equation}\label{eq:tar-q}
\begin{aligned}
q_t&=r(\hat{y}_t;\hat{X}_{1\dots t}, X)\\
&+\sum_{a\in\mathcal{A}}p(a\vert\hat{X}_{1\dots t+1}, \mathbf{c})Q(a;\hat{X}_{1\dots t+1}, X).
\end{aligned}
\end{equation}
The critic network, parameterized by $\theta$, is then updated using the gradient
\begin{equation}\label{eq:q-mse}
\frac{d}{d\theta}\bigg(\sum_{t=1}^{T-1}(Q(\hat{y}_t;\hat{X}_{1\dots t}, X)-q_t)^2\bigg).
\end{equation}

Algorithm~\ref{pc:train} presents the outline of the training process. From Equation~\ref{eq:actor-grad} and Equation~\ref{eq:q-mse}, the training of both the trajectory autoencoder and the critic network, denoted by $p$ and $Q$ respectively, requires the output of the other. This feedback loop can lead to a serious stability issue. To address this, a delayed trajectory autoencoder and a delayed critic network, denoted by $p'$ and $Q'$ respectively, are introduced. Their weights are slowly updated to follow $p$ and $Q$, which are trained in each iteration. We also utilize the $\epsilon$-greedy exploration and the experience replay mechanisms. 

The trajectory autoencoder $p$ is first pre-trained by using maximum log-likelihood, while the critic network $Q$ is pre-trained by using the pre-trained $p$. For each iteration, a trajectory $X$ is sampled by performing a random walk on the pedestrian road network. A sequence of actions $\hat{Y}$ is estimated by feeding $X$ to the encoder and getting predictions from the decoder of the delayed trajectory autoencoder $p'$. An $\epsilon$-greedy exploration is conducted to introduce a small probability of exploring other feasible actions instead of exploiting only the policy generated by $p'$. After getting $\hat{Y}$, the experience is stored in the replay memory.

A batch of $\omega$ experiences is randomly sampled from the replay memory. At each time step $t$, the target $q^k_t$ of the $k$-th experience is computed and used to update the critic network $Q$. We then update the trajectory autoencoder $p$ to maximize the value. Before proceeding to the next iteration, we slowly update the weights of the delayed trajectory autoencoder and the delayed critic network, i.e., $\phi'$ and $\theta'$, by using two constants $\gamma_\phi$ and $\gamma_\theta$. The training process continues until convergence.

\begin{algorithm}[t]
	\caption{Training Trajectory Autoencoder}\label{pc:train}
	\begin{algorithmic}[1]
		\Require A pedestrian road network $G$, a pre-trained trajectory autoencoder $p$ and a pre-trained critic network $Q$ with weights $\phi$ and $\theta$ respectively.
		
		\State Initialize the delayed trajectory autoencoder $p'$ and the delayed critic network $Q'$ with the same weights: $\phi'=\phi, \theta'=\theta$
		\State Initialize the replay memory $\mathfrak{E}=\emptyset$
		\While{not converged}
		\State Receive a trajectory $X$ by random walk on $G$
		\State Generate $\hat{Y}$ from $p'$ with $\epsilon$-greedy exploration
		\State Store experience $(X, \hat{Y})$ in $\mathfrak{E}$
		\If{$\vert\mathfrak{E}\vert\ge\omega$}
		\State Sample a random batch of experiences $\{(X^k, \hat{Y}^k)\}_{k=1}^{\omega}$ from $\mathfrak{E}$ with size $\omega$.
		\State Compute targets for the critic network
		$$
		\begin{aligned}
		q_t^k&=r^k(\hat{y}_t^k;\hat{X}_{1\dots t}^k, X^k)\\
		&+\sum_{a\in\mathcal{A}}p'(a\vert\hat{X}^k_{1\dots t+1}, \mathbf{c}^k)Q'(a;\hat{X}^k_{1\dots t+1}, X^k)
		\end{aligned}
		$$
		\State Update the critic network using the gradient
		$$
		\frac{d}{d\theta}\bigg(\sum_{k=1}^{\omega}\sum_{t=1}^{T-1}\big(Q(\hat{y}^k_t;\hat{X}^k_{1\dots t},X^k)-q_t^k\big)^2\bigg)
		$$
		\State Update the trajectory autoencoder using the gradient
		$$
		\begin{aligned}
		\frac{d}{d\phi}\bigg(&\sum_{k=1}^{\omega}\sum_{t=1}^{T-1}\sum_{a\in\mathcal{A}}p(a\vert\hat{X}^k_{1\dots t},\mathbf{c}^k)Q'(a;\hat{X}^k_{1\dots t},X^k)\bigg)
		\end{aligned}
		$$
		\State Update the delayed trajectory autoencoder and the delayed critic network with constants $\gamma_\phi$ and $\gamma_\theta$
		$$\phi'=\gamma_\phi\phi+(1-\gamma_\phi)\phi',\theta'=\gamma_\theta\theta+(1-\gamma_\theta)\theta'$$
		\EndIf
		\EndWhile
		
	\end{algorithmic}
\end{algorithm}

\section{Experimental Evaluation}
The vector representation of a trajectory should be close, in the feature space, to that of other trajectories with similar spatiotemporal characteristics and far away from those dissimilar. In light of this property, we evaluate our proposed scheme through trajectory clustering.

\subsection{Experimental Settings}
\begin{figure}
\centering
\includegraphics[width=0.7\columnwidth]{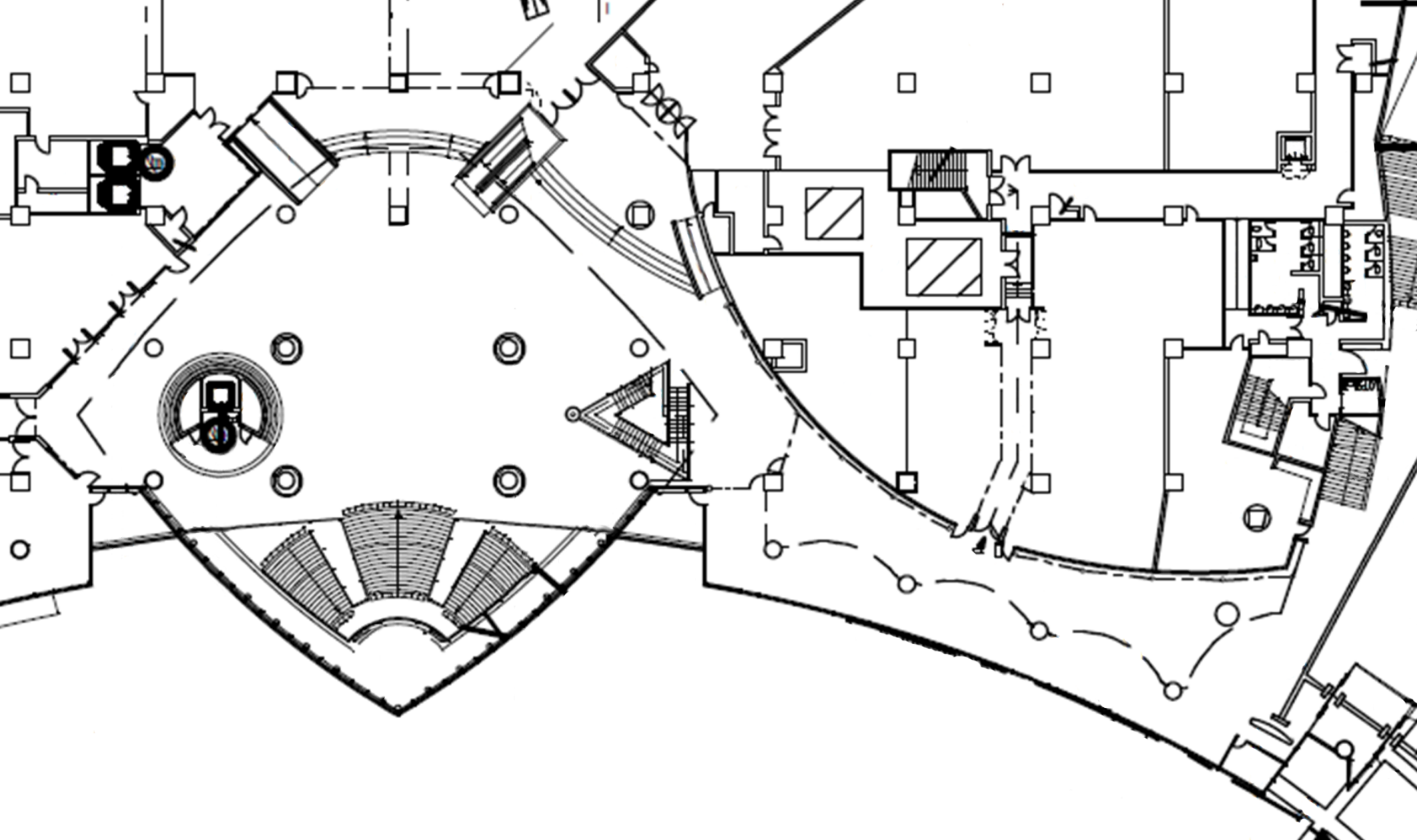}
\caption{The floor plan of the ATC shopping center.}
\label{fig:map-atc}
\end{figure}
An indoor tracking dataset~\cite{brvscic2013person} collected at the Asia \& Pacific Trade Center (ATC) in Osaka, Japan, which includes a shopping center and office buildings, is utilized. Figure~\ref{fig:map-atc} shows the floor plan of ATC where an approximately $900\mbox{m}^2$ area is covered. A set of $5,333$ trajectories, generated by the office workers and travelers, with an average length of $138$ and a sampling interval of $2$ seconds is considered. In addition, a synthetic dataset is carefully generated so that we have six groups of labeled trajectories with different spatiotemporal characteristics for evaluation. 

The baseline parameters are as follows. A grid overlay of size $\alpha=3\mbox{m}$ is applied on the ATC map, and representations of size $D=64$ are generated. The tolerable step-wise reconstruction error $\delta$ is set to $3$, while the critic network has a hidden layer with $512$ neurons. The update parameters $\gamma_\phi$ and $\gamma_\theta$ are both $0.001$.

We compare \schemename{} with three representative approaches:
\begin{itemize}
\item \textbf{DFT}: \emph{Discrete Fourier Transform} extracts the frequencies of the x- and y-coordinates by treating them as two separate signals. The top frequencies in each signal are concatenated to form the vector representation.
\item \textbf{CSSRNN}: \emph{Constrained State Space Recurrent Neural Network}~\cite{wu2017modeling} models trajectories by a standard encoder-decoder architecture with the maximum log-likelihood training. It introduces the concept of masking in the output layer which enforces transition constraints so that the location of the next time step should be directly reachable from the current location.
\item \textbf{\schemename{}-LL}: \emph{\schemename{} using maximum log-likelihood} is a variant of the proposed scheme. It is optimized only by maximizing Equation~\ref{eq:logll} without using the spatial-aware objective function. Different from CSSRNN which models location points directly, \schemename{}-LL predicts the sequence of actions taken by the pedestrian. This scheme shows the power of the spatial-aware objective function.
\end{itemize}
The parameters of the above approaches are selected to produce representations with the same size for comparisons.

We evaluate the quality of the cluster arrangement by using the within cluster sum of error (WCSE). The error between two trajectories $X^i$ and $X^j$, assuming that $\vert X^i\vert<\vert X^j\vert$, is computed by 
\begin{equation}
	error(X^i, X^j) = \mathop{\min}_{0\leq\Delta\leq \vert X^j\vert-\vert X^i\vert}\sum_{t=1}^{\vert X^i\vert}d(x^i_t, x^j_{t+\Delta}).
\end{equation}
The WCSE is then obtained by aggregating the errors between trajectories and the centroid of their assigned cluster. A good cluster arrangement should minimize the WCSE, i.e., all trajectories are close to the centroid.

\subsection{Exploitation of Spatial Dimension}
\begin{figure}
\centering
\includegraphics[width=\columnwidth]{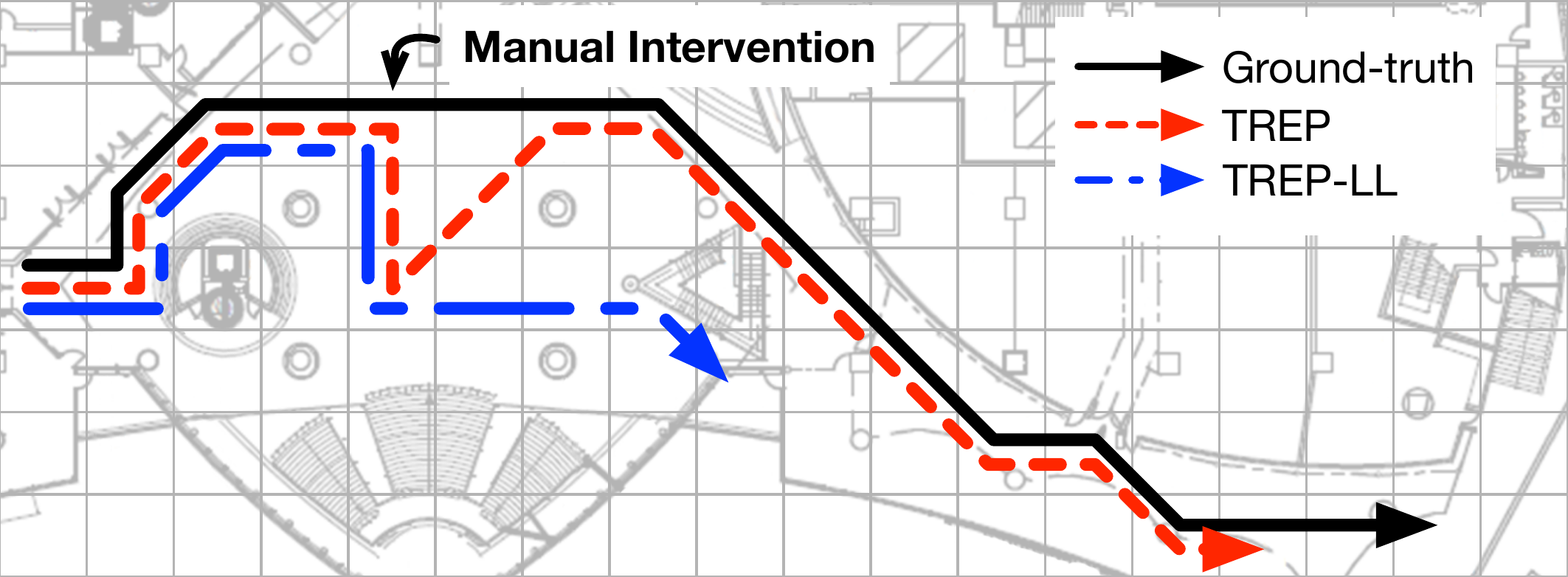}
\caption{Illustration of one ground-truth and two reconstructed trajectories with manual interventions during the decoding phase.}
\label{fig:recoverability}
\end{figure}
To better model trajectories, the trajectory autoencoder should exploit the spatial dimension (e.g., reachability, proximity). Figure~\ref{fig:recoverability} shows one ground-truth and two reconstructed trajectories produced by \schemename{} and \schemename{}-LL. We set $\delta$ to $0$ in this experiment. Originally, both algorithms can reconstruct the exact ground-truth trajectory. We manually assign incorrect actions to make the pedestrians in the decoding phase deviate from the ground-truth. \schemename{} exhibits a thorough understanding of the environment and returns to the correct path within a few moves, since it does not simply memorize the sequence of actions but intelligently produces actions that can maximize the expected future reward. \schemename-LL is incapable of correcting the trajectory as it has not considered the spatial dimension and tends to output the memorized actions.

\subsection{Trajectory Clustering}
\begin{figure}
	\centering
	\begin{subfigure}[b]{0.225\textwidth}
		\includegraphics[width=\textwidth]{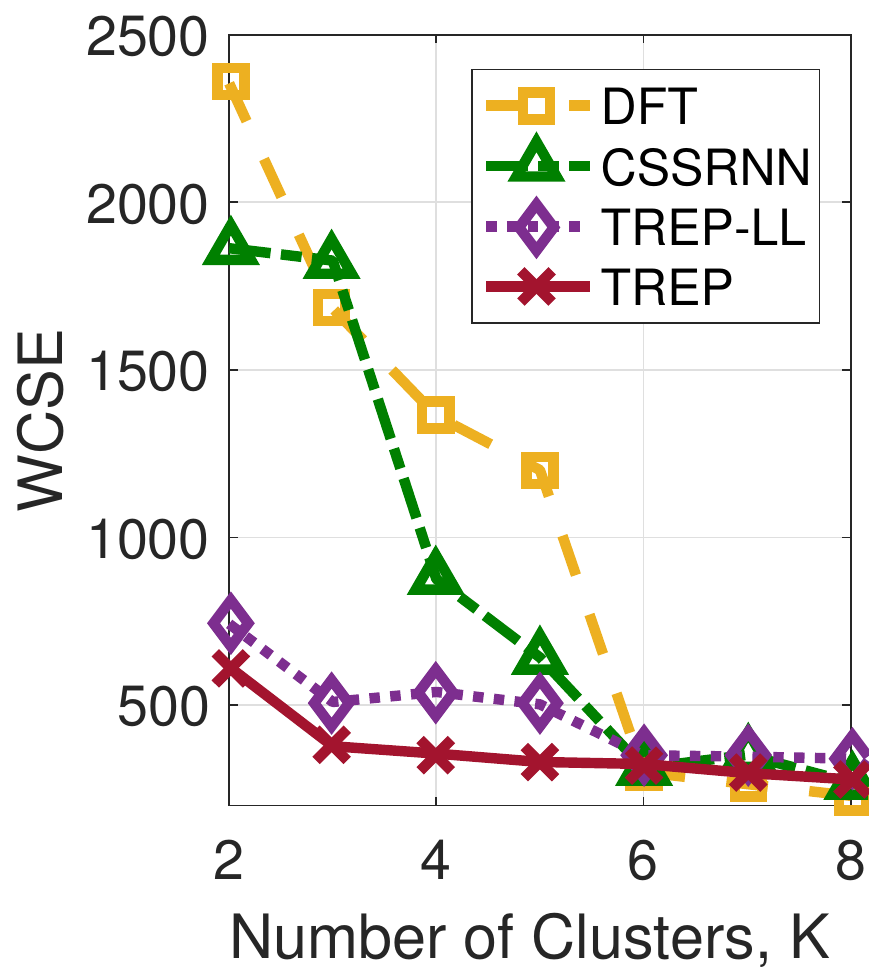}
		\caption{Synthetic dataset}
		\label{fig:cost-syn}
	\end{subfigure}
	\begin{subfigure}[b]{0.225\textwidth}
		\includegraphics[width=\textwidth]{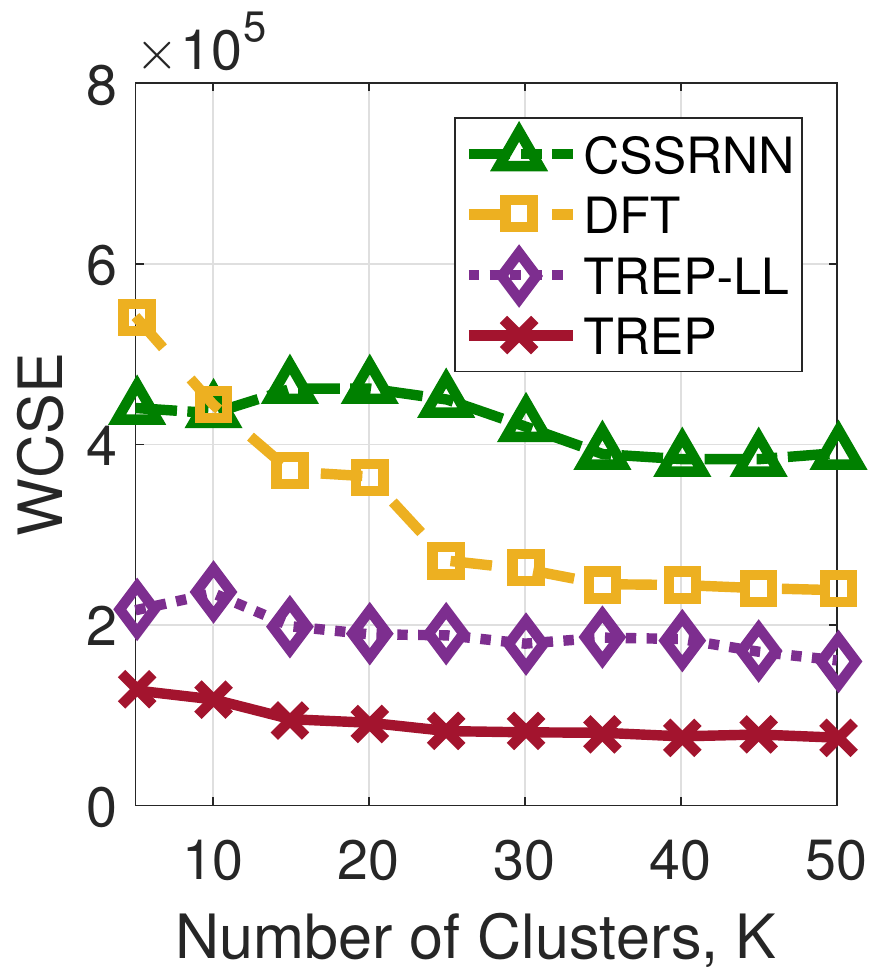}
		\caption{Real dataset}
		\label{fig:cost-real}
	\end{subfigure}
	\caption{WCSE with different schemes.}
	\label{fig:cost}
\end{figure}
Figure~\ref{fig:cost-syn} shows the performance of different schemes on the synthetic dataset. Both schemes can generate high-quality cluster arrangements when the number of clusters is at least six. The ability to summarize and group trajectories with similar spatiotemporal characteristics can be observed when a small number of clusters (i.e., $K<6$ in this case) are formed. \schemename{} and \schemename{}-LL outperform DFT and CSSRNN as the modeling of sequences of actions emphasizes the pedestrian dynamics during the training process, leading to a better summarization ability. CSSRNN is slightly better than DFT as the spatiotemporal characteristics of the trajectories in the synthetic dataset are comparatively simple and can be captured by the neural network.

The performance on the real dataset is shown in Figure~\ref{fig:cost-real}. \schemename{} outperforms other schemes by a large margin. This significant improvement can be attributed to the spatial-aware objective function. Instead of trying to replicate the input trajectory in the decoder using maximum log-likelihood, \schemename{} minimizes the step-wise reconstruction error on the pedestrian road network, which guides the trajectory autoencoder to exploit the spatial dimension. Hence, the ability to handle unseen patterns is significantly better than the others as the real dataset contains trajectories of different lengths and spatiotemporal characteristics, which cannot be enumerated during the training phase. DFT performs better compared with CSSRNN since it is not a data-driven parametric technique. The modeling of location points in CSSRNN cannot be generalized to unseen patterns.

\subsection{Parameter Analysis}
\begin{figure}
	\centering
	\begin{subfigure}[b]{0.22\textwidth}
		\includegraphics[width=\textwidth]{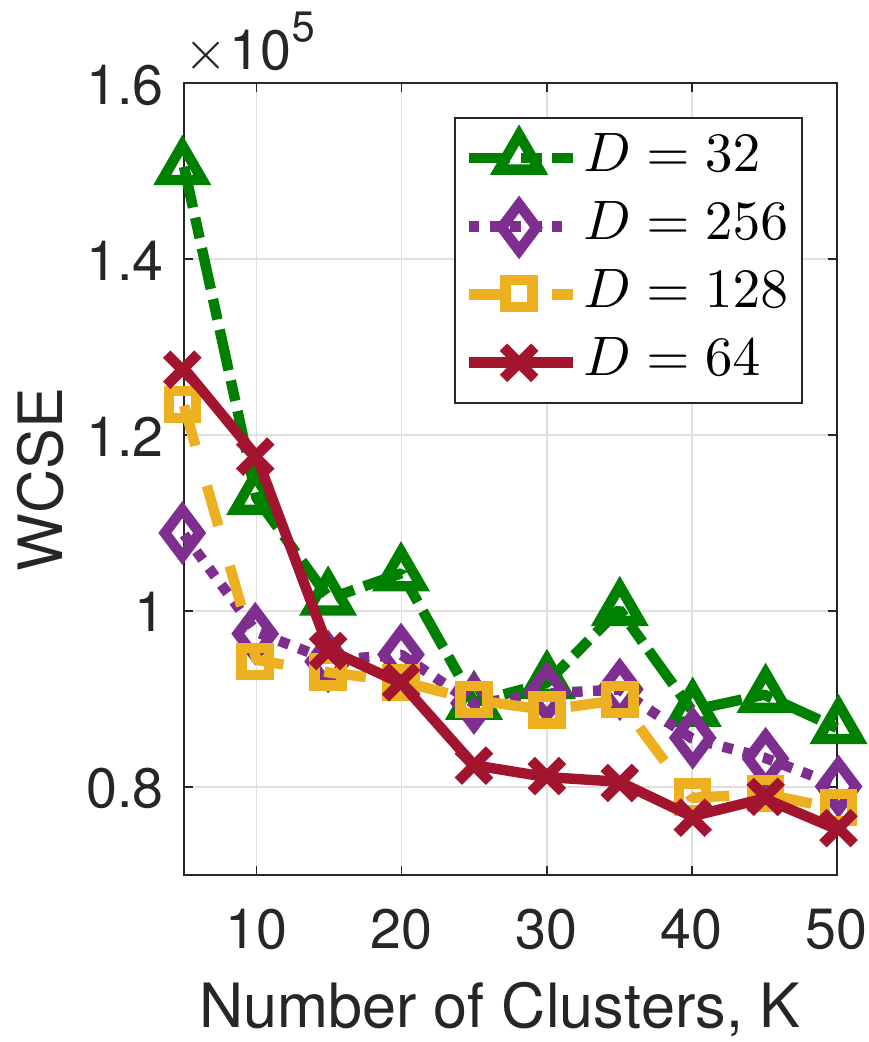}
		\caption{Representation size $D$}
		\label{fig:param-D}
	\end{subfigure}
	~
	\begin{subfigure}[b]{0.22\textwidth}
		\includegraphics[width=\textwidth]{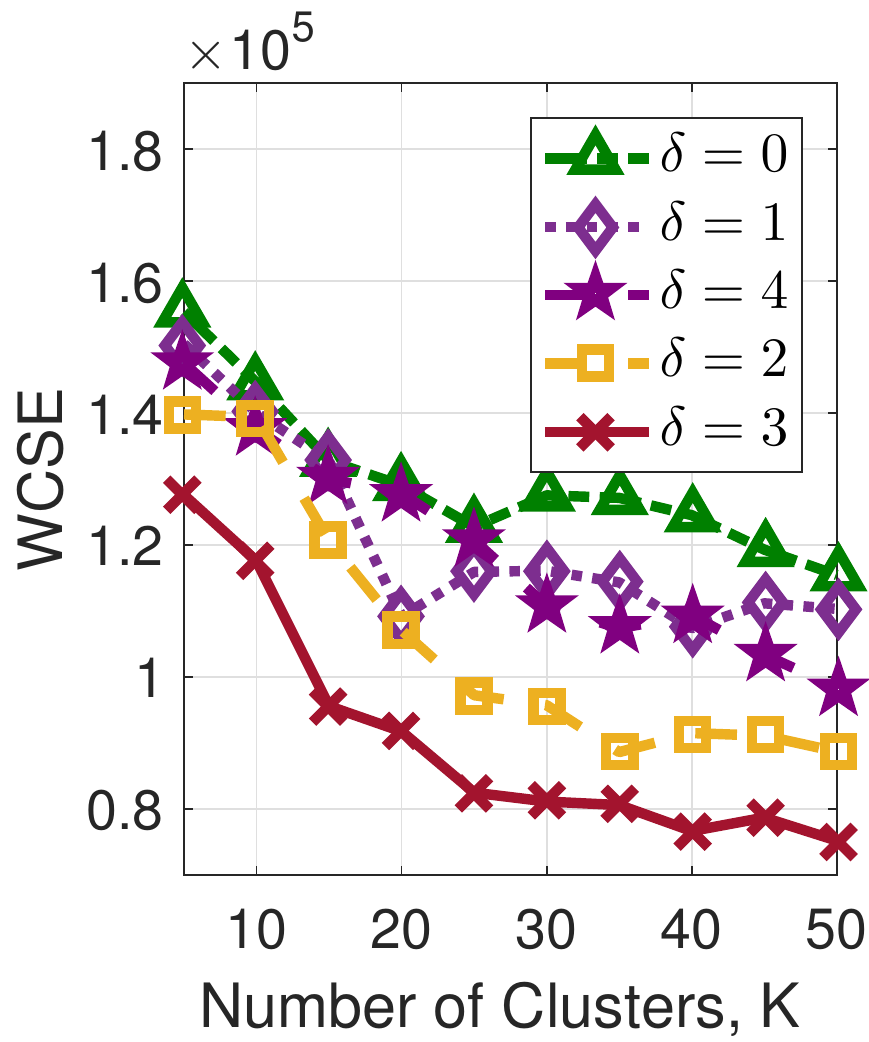}
		\caption{Reconstruction error $\delta$}
		\label{fig:param-delta}
	\end{subfigure}
	\caption{\schemename{} with different parameter settings on the real dataset.}
	\label{fig:param}
\end{figure}
We conduct experiments to observe the performance of \schemename{} with different settings. The studies on two major parameters are presented. Four variants of \schemename{} are built to generate vector representations with sizes $32$, $64$, $128$ and $256$. The results shown in Figure~\ref{fig:param-D} are consistent with our expectation that a larger representation size leads to a better cluster arrangement. However, the improvement is insignificant and can be observed only if a very small number of clusters are formed. Figure~\ref{fig:param-delta} shows the clustering performance with different tolerable step-wise reconstruction errors. Recall that $\delta=0$ means the reconstructed trajectory must be exactly the same as the ground-truth to get a non-negative reward. With a smaller error, we can expect that the trajectory autoencoder focuses on generating an action sequence that can capture every movement of the pedestrian. In this study, we find that \schemename{} exhibits better overall performance when a reasonable amount of error is allowed. The trajectory autoencoder in this setting models not only the local movements but also the global dynamics, which is a crucial property for summarizing trajectories with diverse patterns.

\section{Related Work}
The concept of trajectory representation learning has been the backbone of many exciting data mining applications. \cite{yin2014mining} combines the partial-view features (e.g., velocity and acceleration) and the entire-view features (produced by Discrete Fourier Transform) to form the representations of the trajectories, which are then used to build the user profiles and conduct trajectory recommendations. \cite{yao2017trajectory} extracts motion characteristics (e.g., speed and rate of turn) using a sliding window. Space- and time-invariant features are obtained to perform trajectory clustering. To predict transportation modes given GPS trajectories, \cite{jiang2017trajectorynet} develops representations by handcrafting the semantics (e.g., fast or slow) of the continuous features (e.g., speed).

However, the above works characterize a trajectory by its speed and acceleration patterns without considering the environmental constraints introduced by the road network. To address this, \cite{han2017systematic} proposes a systematic approach to cherry-pick a subset of the training trajectories to form the feature space, while \cite{lee2011mining} constructs the feature space by finding discriminative sequential patterns in the training trajectories. They cannot be generalized well to unseen patterns unless a vast amount of training data covering all the possible spatiotemporal characteristics is obtained.

Learning representations using end-to-end neural networks has been a promising approach. \cite{gao2017identifying} and \cite{zhou2018trajectory} represent trajectories as embeddings by using a recurrent neural network or a variational autoencoder. To analyze driving behaviors, \cite{wang2018you} first derives the driving operation transition graph from the GPS trajectories. An autoencoder is then used to obtain the vector representations. They cannot be extended to pedestrian trajectories as environmental constraints are not considered. \cite{endo2016classifying} converts a trajectory into an image where the pixel intensity represents the dwell time of the corresponding location. A deep neural network is then utilized to learn the hidden representations. Though dwell time can be considered, the ordering information is lost. \cite{wu2017modeling} uses a recurrent neural network with a state-constrained softmax in the output layer to model trajectories, while \cite{lv2018lc} proposes topology-aware look-up operations to exploit the nearby traffic conditions in a traffic speed prediction scenario. With the above neural network architectures, a trajectory autoencoder considering both spatial and temporal dimensions can be built. However, an effective learning objective guiding the autoencoder to understand the environment and the pedestrian dynamics is necessary to train a model capable of handling diverse trajectory patterns.

\section{Conclusion}
We have studied the problem of  representation learning for pedestrian trajectories. We propose a novel framework, \schemename{}, with a spatial-aware objective function trained under the paradigm of actor-critic reinforcement learning. \schemename{} bridges the gap between trajectory data and feature-based data mining. The fixed-length vector representation extracted from a variable-length trajectory encodes the salient features compactly. Extensive experimental evaluations on both synthetic and real datasets from the ATC shopping center show that \schemename{} effectively encodes spatiotemporal characteristics of trajectories with a high fidelity. In the future, we will generalize the proposed framework to handle irregularly sampled and noisy trajectories.

\bibliographystyle{aaai}

\end{document}